# The Effect of Gender Diversity on Scientific Team Impact: A Team Roles Perspective

Yi Zhao[1], Yongjun Zhu[2,3], Donghun Kim[4], Yuzhuo Wang[1], Heng Zhang[5], Chao Lu[6], Chengzhi Zhang[7*]

1. School of Management, Anhui University, Hefei, China
2. Department of Library and Information Science, Yonsei University, Seoul, Republic of Korea
3. Center for Science of Science, Yonsei University, Seoul, Republic of Korea
4. School of Information Management, Nanjing University, Nanjing, China
5. School of Information Management, Central China Normal University, Wuhan, China
6. Business School, Hohai University, Nanjing, China
7. Department of Information Management, Nanjing University of Science and Technology, Nanjing, China

**Abstract:** The influence of gender diversity on the success of scientific teams is of great interest to academia. However, prior findings remain inconsistent, and most studies operationalize diversity in aggregate terms, overlooking internal role differentiation. This limitation obscures a more nuanced understanding of how gender diversity shapes team impact. In particular, the effect of gender diversity across different team roles remains poorly understood. To this end, we define a scientific team as all coauthors of a paper and measure team impact through five-year citation counts. Using author contribution statements, we classified members into leadership and support roles. Drawing on more than 130,000 papers from PLOS journals, most of which are in biomedical-related disciplines, we employed multivariable regression to examine the association between gender diversity in these roles and team impact. Furthermore, we apply a threshold regression model to investigate how team size moderates this relationship. The results show that (1) the relationship between gender diversity and team impact follows an inverted U-shape for both leadership and support groups; (2) teams with an all-female leadership group and an all-male support group achieve higher impact than other team types. Interestingly, (3) the effect of leadership-group gender diversity is significantly negative for small teams but becomes positive and statistically insignificant in large teams. In contrast, the estimates for support-group gender diversity remain significant and positive, regardless of team size.

---

[*] Corresponding author

*Email addresses*: yizhao93@ahu.edu.cn (Yi Zhao), zhu@yonsei.ac.kr (Yongjun Zhu), dhkim91@yonsei.ac.kr (Donghun Kim), wangyuzhuo@ahu.edu.cn (Yuzhuo Wang), zh_heng@ccnu.edu.cn (Heng Zhang), luchao91@hhu.edu.cn (Chao Lu), zhangcz@njust.edu.cn (Chengzhi Zhang)

# 1. Introduction

A comprehensive examination of teamwork cannot overlook the significant role of gender. From the perspective of social construction, gender is a social identity that affects thoughts, perception, and worldview (Lindsey 2015). Gendered aspects of this social identity can influence researchers' motivational dispositions and cognitive processes (McLeod 2005). Consequently, this guides researchers' behaviors, actions and perceptions. Previous studies have shown that gender differences manifest across many aspects of scientific activities, including research methods and topic selection (Zhang et al. 2023; Thelwall et al. 2019; Wullum Nielsen & Börjeson 2019), collaborative preferences (Boschini & Sjögren 2007), and the division of labor in scientific teamwork (Larivière et al. 2020). For example, in the field of tourism, Nunkoo et al. (2020) indicated that male researchers more likely to adopt quantitative methods to address problems, whereas female researchers tend to prefer qualitative approaches. Based on self-reported contribution statements in publications, Larivière et al. (2020) further showed that male scholars are more often engaged in conceptual work in teams, while female scholars more frequently conduct empirical tasks. When male and female scholars collaborate, their gender-differentiated beliefs, experiences, and perspectives can be combined to enhance team performance and efficiency (Nielsen et al. 2018).

Over the past few decades, the increased participation of women in science has promoted the gender diversity of scientific teams (Huang et al. 2020). A recent study by Yang et al. (2022) revealed that the proportion of gender-heterogeneous teams in medicine increased from 60% to 70% between 2000 and 2019. The impact of gender diversity on the performance of scientific teams has garnered extensive scholarly attention. Different gender categories may offer distinctive worldviews, perspectives and cognitive approaches (Wullum Nielsen & Börjeson 2019). Compared with gender-homogenous teams, gender-diverse teams have greater potential to pool a wider range of knowledge, beliefs, and experiences, thereby fostering innovation and creativity (Campbell et al. 2013). Nevertheless, some studies reached the opposite conclusion (Joshi 2014; Stvilia et al. 2011). For instance, Wullum Nielsen and Börjeson (2019) analyzed more than 25,000 article teams in management and found that gender diversity has no significant effect on team outcomes. These mixed findings suggest that further investigation into the potential association between gender diversity and team performance is necessary.

Although a few scholars still conduct independent research without collaboration, teamwork has become the dominant mode of producing scientific innovation in contemporary science (Wuchty et al. 2007). This shift can be attributed to several factors, including the increasing complexity of research questions (Zhao et al. 2022), the growing specialization of scientists (Jones 2009), reduced travel costs (Catalini et al. 2020), and advances in communication technology (Katz & Martin 1997; Naik et al. 2023). Collectively, these factors have enable teams not only to thrive but also to expand in size (Liu et al. 2020). The ever-expanding team size in science has fostered a more specialized division of labor. To our knowledge, most studies on the association between gender diversity and scientific performance examine teams as a whole, often overlooking the roles undertaken by individual members. Yet research suggests that diverse team compositions can enhance the generation of original and valuable ideas, these benefits may diminish during idea selection and implementation (Stahl et al. 2009). Because different roles within scientific teams involve different research tasks, it is important to consider how gender diversity within specific roles influences team impact. Furthermore, although prior studies have recognized team size as a vital factor affecting team innovation (Wu et al. 2019; Milojevic 2014), it is still unclear whether team size moderates the relationship between role-specific gender diversity and team impact. To fill these research gaps, we propose two questions in this study:

*RQ1*: How does gender diversity within different team roles affect the team's impact?
*RQ2*: Can team size moderate this relationship?

The contributions of this paper are primarily manifested in three aspects. First, we unpack the distinct effect of gender diversity in two specific roles (i.e., leadership and support roles) on team impact. Second, we identify team size as a key moderator in the association between gender diversity in leadership groups and team impact. Third, our findings offer valuable insights for practitioners and team leaders regarding the construction of high-impact teams.

## 2. Literature review

In this section, we review four streams of literature relevant to our study. First, we discuss the role of gender homophily in research team formation. Second, we review the literature on scientific teams and team roles. Third, we summarize existing studies on the relationship between role-specific gender diversity and scientific team performance. Finally, we review research on how team size may moderate the relationship between gender diversity and team impact.

### 2.1 Role of gender homophily in research team formation

Homophily principle suggests similarity serves as a basis for interpersonal connection (McPherson et al. 2001). People are naturally attracted to those who resemble them in terms of background, interests, worldview, and beliefs, because similarity is often perceived as reflecting one's ideal self (Byrne et al. 1986). As a result, personal networks tend to be homogeneous with respect to sociodemographic and personal traits such as gender, wealth, education, and age (Kwiek & Roszka 2021).

The formation of research teams follows similar mechanisms. In academia, team formation is a autonomy process, in which scholars decide whether to collaborate based on perceived benefits and costs (McDowell & Smith 2007). Collaboration is pursued when it is expected to yield greater benefits than working alone. Consequently, the composition of research teams reflects individual choices, shaped by anticipated gains from collaboration and the coordination costs involved (Boschini & Sjögren 2007).

Gender homophily within teams may facilitate smoother communication but can also restrict scholars' worldviews, values, and knowledge to a narrower domain compared with gender-diverse teams. As a result, many male scholars tend to collaborate with other men, while many female scholars prefer to collaborate with other women(Prakash et al. 2024). At the same time, some scholars actively seek collaborations with colleagues of the opposite gender. A large-scale study by Kwiek and Roszka (2021) found that female scholars are more likely to participate in mixed-gender teams, while male scholars tend to work in all-male teams.

## 2.2 Scientific teams and team roles

Teamwork is becoming more common and has become the primary way to drive the production of new knowledge in modern science (Wu et al. 2019). The division of labor in scientific research is becoming progressively more refined, which is reflected in the increasingly differentiated roles within research teams (Melero & Palomeras 2015; Hamilton et al. 2003; Lee et al. 2015). Based on the author-task bipartite network, Lu et al. (2020) divided team roles into three types: specialists, team-players and versatiles. Th authors found that versatiles are associated with funding and supervision. Similarly, Robinson-Garcia et al. (2020) adopted archetypal analysis to categorize scientists into distinctive roles and suggested that junior scholars are typically involved in specialized and supporting roles, while late-career scholars are often engaged in leadership and support roles. Utilizing the authorship contribution statement, Zhao et al. (2024) identified the authors who undertake conceiving ideas as thought leaders in scientific teams.

The order of authors in a byline has often been used to infer team roles in science-of-science studies. Despite disciplinary differences in authorship norms, most studies suggest that the first and last authors contribute more substantially than middle authors and can therefore be regarded as occupying leadership roles (van Leeuwen 2008). However, ghost authorship and honorary authorship can substantially bias such inferences (Pallotti et al. 2025). Scholars at different career ages may possess different leadership skills (Mumford et al. 2007). Xu et al. (2024) thus defined team leaders based on career age. Nevertheless, the accuracy of career-age measures in scientometrics relies on effective name disambiguation, and the disambiguation of mainstream bibliometric data remains a significant challenge, particularly for East Asian scholars (Kim et al. 2021). Academic rank may also serve as a useful indicator of leadership roles, as full professors often assume leadership positions in academic teams (Evans et al. 2013). However, large-scale data on academic ranks are difficult to obtain. Therefore, in this study, we relied on author contribution statements to determine roles in scientific teams.

## 2.3 Gender diversity of different team roles and scientific team performance

The literature on gender diversity and its influence on academic team performance from the perspective of division of labor remains scarce, and prior findings on this relationship are inconsistent. Some studies argued that team performance benefits from gender diversity (Maddi & Gingras 2021; De Saá-Pérez et al. 2017). For example, Campbell et al. (2013) revealed that articles published by mix-gender teams received 34% more citations than those authored by same-gender teams. Yang et al. (2022) found that papers produced by gender-heterogeneous teams are significantly more impactful and novel than those produced by gender-homogeneous teams. However, other empirical work suggests that gender diversity have no discernible effect on team performance (Wullum Nielsen & Börjeson 2019). For instance, Stvilia et al. (2011) explored the influence of team diversity and network characteristics on team productivity, discovering that gender diversity had a negative but non-significant effect. Joshi (2014) analyzed 60 teams across science and engineering, and reached a similar conclusion to those of Stvilia et al. (2011).

A potential explanation for these inconsistent findings lies in the gender distribution across disciplines, which constrains the possible gender composition of teams. For example, Ghiasi et al. (2015) reported that female scholars account for only 20% of authorship in Engineering. Using publications indexed in the Web of Science between 1900 and 2016, Huang et al. (2020) found that the proportion of female scholars was as low as 15% in Mathematics and Physics, but reached approximately 30% in Health science, Political science, and Phycology. Similar disadvantages have been observed in other fields. Zhang et al. (2025) showed that woman accounted for only 22.2% of authorship in the Natural Language Processing. Based on more than 6.6 million publications in the medical sciences, Yang et al. (2022) showed that female's participation increased from 38% to 46% between 2000 and 2019, but gender-heterogeneous teams remain underrepresented relative to expectations. Different disciplines exhibit distinct patterns in women's participation and the gender composition of teams. Therefore, further research is needed to investigate the potential influence of team gender diversity on the performance of scientific teams.

In the field of science of team science, there is a limited body of quantitative literature examining the relationship between gender diversity in different team roles and team impact. Studies in other domains have yielded mixed results, yet these studies provide valuable insights. A growing number of literature suggests that team performance benefits from gender diversity in leadership roles (Saeed et al. 2023). For example, Lyngsie and Foss (2016) found a positive relationship between gender diversity in leadership group (i.e., top management groups) and innovation outcomes in established firms, although this positive effect is weakened in firms with a high proportion of female employees. Similarly, Lafuente and Vaillant (2019) used data from all financial firms in Costa Rican industry, highlighted a positive association between gender balance in board configurations and economic performance.

According to information processing theory, diverse teams have a significant advantage in addressing complex problems because they are more likely to access a broad range of knowledge (Phillips et al. 2013). In contrast, other scholars argue that gender diversity in leadership can be detrimental to team performance (Andres et al. 2005). For instance, Lee and Chung (2022) investigated leadership-group gender diversity on firm innovation and found that greater gender diversity reduced teams' innovation impact. According to social categorization theory, leaders are inclined to classify themselves into subgroups based on demographic characteristics, which may foster cognitive biases and intra-team conflict (Phillips & O'Reilly 1998). Consequently, gender-diverse leadership teams may experience low performance compared with more homogeneous ones.

Scientific teams differ substantially from business teams. The primary goal of scientific teams is to solve complex problems and generate highly innovative knowledge, and their work is usually knowledge intensive. By contrast, business teams typically aim to achieve financial performance, and their work is more diverse, encompassing mental labor, physical labor, and hybrid tasks (Börner et al. 2010; Hoogendoorn et al. 2013). In addition, scientific teams have several distinctive characteristics compared with business teams (Whitley 2000). Scientific collaborations are largely voluntary and grounded in mutual interests, and scholars have considerable autonomy to create, maintain, restructure, or dissolve their teams. In contrast, business teams are generally more structured and goal oriented, with leadership and decision-making authority often centralized. Members of business teams may have limited autonomy in forming or dissolving teams (Wang & Hicks 2015; Devine et al. 1999). Moreover, scientific collaboration tends to be fluid and evolve with the project, while long-term partnerships are relatively rare. (Zhu et al. 2024; Wang & Hicks 2015). Business teams, particularly top management teams, are comparatively stable (Liao et al. 2023). Given these differences, further investigation into the influence of gender diversity on team impact in academia becomes particularly necessary.

Taken together, current empirical evidence on the relationship between team gender diversity and team performance in academia is equivocal. The existing literature largely operationalizes diversity in aggregate manners, focusing on overall gender diversity and overlooking the critical role-based mechanisms. Furthermore, although gender diversity within leadership groups (e.g., top management teams) has been recognized as important in the business sector, it remains unclear how gender diversity across different team roles in scientific field relates to team impact.

## 2.4   Team size and team performance

Large teams have both advantages and drawbacks for team performance. A stream of studies has shown that larger and more diverse teams are more likely to produce innovative and impactful outcomes (Wu et al. 2019; Yang et al. 2022). By drawing on a broader pool of resources, large teams increase their chance of developing new and useful ideas (Stewart 2016; Krammer & Dahlin 2023). They also tend to receive more citations from the scientific community (Wang et al. 2017). One possible reason is that larger team can more effectively disseminate their research outputs (Krammer & Dahlin 2023). However, other literature emphasizes the potential "process loss" associated with large teams, which may negatively affect team performance (Cummings et al. 2013). In particular, managing and coordinating the wide range of resources required by large teams can be challenging (Liu et al. 2022). Additionally, larger teams may be more prone to free-riding (He 2012), lower team cohesion (Manners 1975), and increased emotional conflict (Amason & Sapienza 1997).

Team size may moderate the relationship between gender diversity in different team roles and team performance. The upper echelons theory posits that the characteristics of top management teams play a crucial role in decision-making (Hambrick & Mason 1984). Similarly, the characteristics of team leaders in science are critical for shaping scientific outputs. Existing literature suggests that team diversity tends to enhance team performance on complex problems rather than simple ones (Chatman et al. 2019). In small teams, team structure are typically simpler and flatter than in large teams (Xu, Wu, et al. 2022). If leadership groups in small teams experience intergroup bias and subgroup categorization, with leaders aligning strongly with their respective ingroups, this may lead to conflict, distrust, and dislike among subgroup leaders (Homan et al. 2020). Such processes can undermine overall team performance. In contrast, larger teams tend to be more diverse (Yang et al. 2022), and diverse leadership groups may have greater capacity to manage and lead such teams, which can facilitate team innovation (van Knippenberg & Schippers 2007; Homan et al. 2020). Therefore, we speculate that the positive effect of leadership diversity is likely to be less pronounced in small teams but more significant in large teams.

While numerous studies have examined how team size influences team performance, little is known about whether it moderates relationship between role-specific gender diversity and team impact.

# 3 Data

In this section, the detailed data collection and data processing approach will be introduced.

## 3.1 Data collection

The dataset used in this study was obtained from PLOS, which is a prominent open-access publisher in the scientific community. According to PLOS publisher's publication policy[1], each author's contributions to the study should be described in the article. This policy provides an opportunity to explore the influence of gender diversity on scientific team performance from the perspective of team roles. To acquire the author contribution statements and other metadata, including affiliations and author names, we collected 161,193 full-text research articles from PLOS journals published between 2007 and 2015 in JATS-standard XML format[2]. Due to inconsistencies in the taxonomy of contribution statements before and after 2016 (Lu et al. 2020), our analysis was restricted to articles published between 2007 and 2015.

---
[1] https://journals.plos.org/plosone/s/authorship

## 3.2 Data processing

Figure 1 presents an illustrative example and overall flowchart of data collection and processing. Figure 1A provides a representative example of the data processing procedure, which can be divided into four main components: (1) author-task pairs extraction, (2) author gender inference, (3) team roles assignment, and (4) discipline assignment. Figure 1B outlines the detailed steps involved in dataset construction.

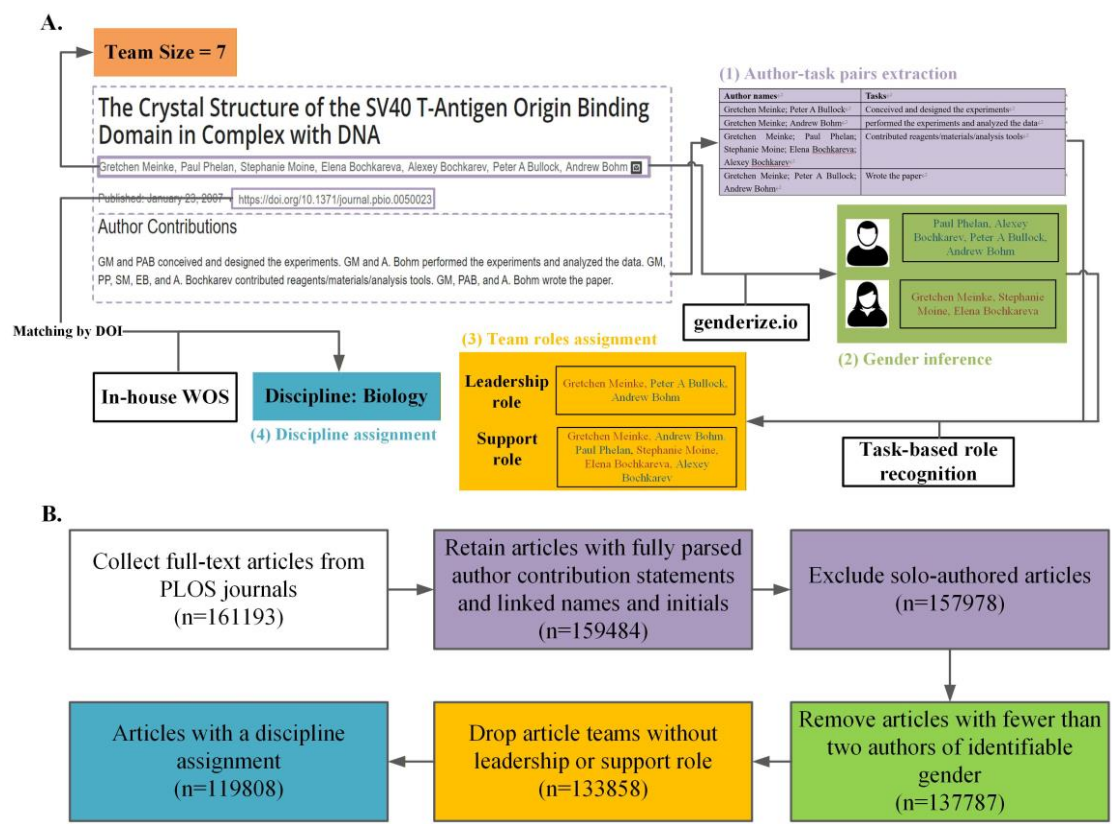

Figure 1 Illustrative example and flowchart of data collection and processing

**(1) Author-task pairs extraction**

Figure 2A shows an example of author contribution statement in JATS-standard XML format. The author-task pairs extraction proceeded in a two-step process. First, we adopted regular expressions to parse the author contribution statement embedded between the start tag <fn fn-type="con" id="ack1"> and the end tag </fn>. The parsed results are presented in Figure 2B. Second, we utilized string matching to match the authors' full names in the authorship with their initials from the author-task pairs (see

---
[2] https://jats.nlm.nih.gov/publishing/

Figure 2C). We retained only those articles for which all authors' contribution statements were fully parsed, and where their full names and initials were successfully linked, yielding 159,484 articles. Since the focus of this study is on team-level analysis, we further excluded 1,506 solo-authored articles, leaving 157,978 articles for analysis.

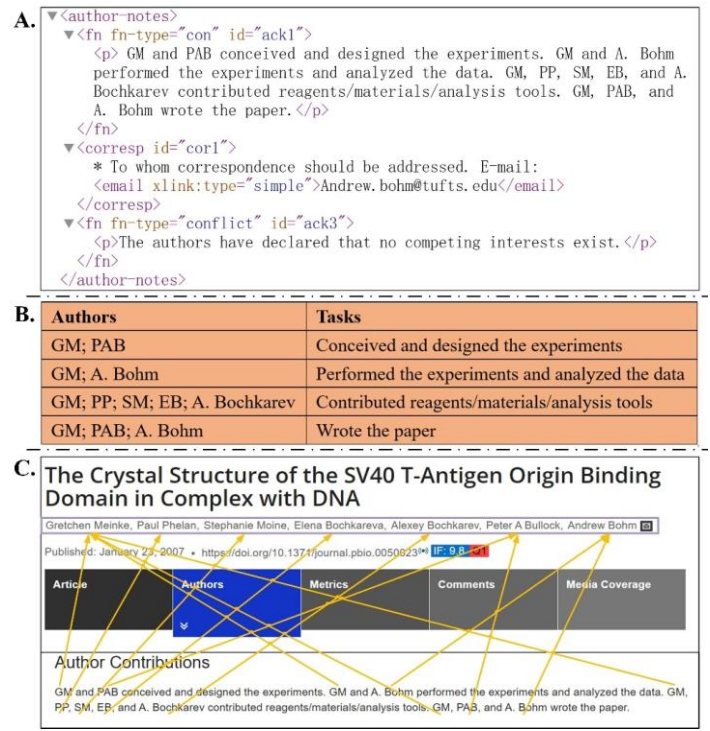

Figure 2 The author–task pairs of a sample article[3]

**(2) Author gender inference**

Name-based gender inference has been widely applied in studies of team science. In this study, we used Genderize.io[4] to infer gender for each scholar based on their first name (Madsen et al. 2022). Huang et al. (2020) reported an overall accuracy of 94.25%, suggesting that the tool is reasonably reliable. However, it should be noted that the Genderize.io is limited to binary gender categories and is not intended to address the issue of nonbinary gender identities. Using this tool, 93.05% of authors were assigned a gender, as shown in the gender inference statistics in Table 1. Another limitation of Genderize.io is its relatively poor performance with Chinese and Korean first names. Following widely accepted practice (Huang et al. 2020), we excluded the authors with 60 most common Chinese[5] and Korean[6] surnames. After gender inference, authors whose gender could not be determined were excluded from the dataset. Specifically, if an article had three authors, two identified as male and one with undetermined gender, the article was retained, but the undetermined author was excluded from the analysis (Maddi & Gingras 2021), leaving 149,587 articles. Articles were preserved only if the gender of at least two authors could be inferred. As a result, the dataset consisted of 137,787 articles.

---

[3] https://journals.plos.org/plosbiology/article?id=10.1371/journal.pbio.0050023
[4] https://genderize.io/our-data
[5] http://en.wikipedia.org/wiki/List_of_common_Chinese_surnames
[6] http://en.wikipedia.org/wiki/List_of_Korean_family_names

Table 1 Gender distribution of authors

| Inferred gender | Percentage (%) | Number of distinct authors |
|---|---|---|
| Male | 55.00% | 343,033 |
| Female | 38.05% | 237,356 |
| Unknown | 6.95% | 43,384 |

**(3) Team roles assignment**

Drawing on self-reported contribution statements, Xu, Wu, et al. (2022) categorized authors into different team roles. Specifically, they adopted the Louvain method to cluster the co-occurrence network of tasks into three groups: leadership, direct support and indirect support. The mapping scheme between team roles and tasks is summarized in Table 2.

Table 2 Mapping between team roles and tasks

| No. | Team roles | Tasks |
|---|---|---|
| 1 | Leadership | "conceive", "supervise", "design", "lead", "coordinate", "write", and "interpret". |
| 2 | Direct support | "help", "assist", "prepare", "develop", "collect", "generate", "purify", "carry", "do", "perform", "conduct" and "analyze". |
| 3 | Indirect support | "participate", "provide", "contribute", "comment", "discuss" and "edit". |

According to the mapping scheme in Table 2, each author could be assigned to one or more of the three roles based on their reported tasks. Using this scheme, approximately 98% of authors were assigned at least one role. Because the mapping scheme was developed from the 25 most common tasks (see Table 2), authors performing tasks outside this set could not be directly mapped to any of the three roles. To address this limitation, we designed a supplementary strategy. Specifically, we first converted all tasks into a 384-dimensional vector using a pre-trained model[7]. We then calculated the cosine similarity between the unlabeled tasks and the 25 labeled tasks (Salton & McGill 1983). Finally, each unlabeled task was assigned the role of the most similar labeled task. In practice, the distinction between direct support and indirect support is often ambiguous, we thus merge these categories into a single support role. Team members could simultaneously serve in leadership and support roles, in which case they were classified into both groups. Since our study focuses on the roles of leadership and support in shaping team impact, 3,929 article teams lacking either role were excluded from the dataset. After this step, the dataset comprised 133,858 articles.

**(4) Discipline assignment**

To control for discipline-invariant factors in the regression analysis, it was necessary to assign a discipline to each PLOS article. We utilized discipline classification information from the in-house Web of Science (WoS) database, owned by

---
[7] https://huggingface.co/sentence-transformers/all-MiniLM-L6-v2

Dr. Vincent Larivière, to determine the discipline of each article in our dataset, as done in our previous study (Lu et al. 2019). The classification system used for this dataset is derived from the subject classification system of the National Science Foundation[8]. The specific method employed involved using the category to which the majority of the references from PLOS articles belonged as the discipline label for these articles. For instance, suppose article A has 40 references, and the distribution of discipline categories from these 40 references' journals is counted as $\{S_1, S_2, ...\}$, the discipline $S_i$ containing the most references was identified as the discipline label for article A. After matching, 119,808 articles were assigned to a discipline. Table 3 presents the disciplinary distribution of our dataset. To retain 14,049 the articles without discipline assignments and maintain a larger dataset, we categorized them under a single group, labeled "other discipline", in the subsequent analysis. Consequently, the final dataset comprised 133,858 articles.

Table 3 Number of publications by discipline

| Discipline | # of publications | Ratio (%) |
|---|---|---|
| Clinical Medicine | 54,020 | 40.36 |
| Biomedical Research | 43,218 | 32.29 |
| Unmatched | 14,049 | 10.50 |
| Biology | 11,293 | 8.44 |
| Mathematics | 2,909 | 2.17 |
| Psychology | 2,809 | 2.09 |
| Earth and Space | 1,332 | 1.00 |
| Health | 1,038 | 0.77 |
| Social Sciences | 812 | 0.61 |
| Physics | 811 | 0.61 |
| Engineering and Technology | 715 | 0.53 |
| Chemistry | 488 | 0.36 |
| Professional Fields | 349 | 0.26 |
| Humanities | 14 | 0.01 |
| Arts | 1 | 0.00 |
| Total | 133,858 | 100 |

We also present the distribution of publications by journal, as shown in Table 4. It can be observed that 86.90% of the articles were published in *PLOS One*, whereas only 13.1% articles were from the other seven journals.

Table 4 Distribution of publications across journals

| Journal | # (Ratio) | Journal | # (Ratio) |
|---|---|---|---|
| PLOS One | 116,324 (86.90%) | PLOS Neglected Tropical Diseases | 3,196 (2.39%) |
| PLOS Genetics | 4,558 (3.41%) | PLOS Biology | 1,443 (1.08%) |
| PLOS Pathogens | 3,958 (2.96%) | PLOS Medicine | 1095 (0.82%) |

---
[8] http://www.nsf.gov/statistics/nsf13327/pdf/tabb1.pdf

| PLOS Computational Biology | 3265 (2.44%) | PLOS Clinical Trials | 19 (0.01%) |

# 4 Methodology

In this section, we first introduce measures of team impact and team gender diversity, and then describe the two regression methods employed in the analysis.

## 4.1 Measuring team impact

Citation counts are widely used as a proxy for research team impact in the science-of-science studies. Following the recommendation of Wang (2012), and consistent with common practice in team science studies (Xu et al. 2024), we use a five-year citation window to measure team impact..

To obtain annual citation counts of the articles, we linked the PLOS dataset with the PubMed Knowledge Graph, which is an enhanced version of PubMed that integrates multiple sources of citation data, including PubMed's own citation data, NIH's open citation collection, OpenCitations, and citation data from WoS (Xu et al. 2020). Because the PLOS dataset identifies publications by DOI, whereas the PubMed Knowledge Graph primarily uses PMID, we first retrieved the PMID for PLOS articles using the ID Converter (https://www.ncbi.nlm.nih.gov/pmc/tools/idconv/). Each article was then matched based on its PMID.

## 4.2 Measuring team gender diversity

To quantify the degree of gender diversity in different team roles, we adopt two types of measures: a continuous measure and a binary measure. For the continuous measure (Yang et al. 2022), the specific formula is given as follows:

$$TGD_{ij} = -p_{ij}log_2(p_{ij}) - (1-p_{ij})log_2(1-p_{ij}), j \in \{1,2,3\} \quad (1)$$

Where $TGD_{ij}$ represents team gender diversity of type $j$ in team $i$, and $p_{ij}$ denotes the proportion of female authors of type $j$ in team $i$. When $j = 1$, it refers to the whole team; when $j = 2$, it refers to the leadership-group gender diversity; and when $j = 3$, it refers to the support-group gender diversity. The range of $TGD$ is from 0 to 1. When $TGD = 0$, it denotes the team consists of all males or all females. When $TGD = 1$, it means the proportions of males and females in the team are equal. When $0 < TGD < 1$, it implies that either males or females constitute the majority in the team. For example, suppose a team consists of two males and one female, and the proportion of female authors is 0.333. In this case, the $TGD$ would be 0.918.

For the binary measure, a team composed of both male and female members is classified as gender-heterogeneous, whereas a team composed of only one gender is classified as gender-homogeneous. We further categorized teams by the gender composition of their leadership and support groups into nine types, as shown in Table 5. In the subsequent regression analysis (Table 7, Model 8), team type is represented by a set of dummy variables, and XLead-XSup serves as the reference category.

Table 5 Team types by gender composition of leadership and support roles

| No. | Team types | Abbreviation |
|---|---|---|
| 1 | Mixed-gender leadership group & Mixed-gender support group | XLead-XSup |
| 2 | Mixed-gender leadership group & Male-only support group | XLead-MSup |
| 3 | Mixed-gender leadership group & Female-only support group | XLead-FSup |
| 4 | Male-only leadership group & Mixed-gender support group | MLead-XSup |
| 5 | Female-only leadership group & Mixed-gender support group | FLead-XSup |
| 6 | Male-only leadership group & Male-only support group | MLead-MSup |
| 7 | Male-only leadership group & Female-only support group | MLead-FSup |
| 8 | Female-only leadership group & Female-only support group | FLead-FSup |
| 9 | Female-only leadership group & Male-only support group | FLead-MSup |

## 4.3 Multivariable and threshold regression

**(1) Multivariable regression**

Multivariable regression was used to address RQ1. It is well known that citations can be influenced by various factors. According to previous studies, we included team size (Yang et al. 2022), international team (Kuan et al. 2024), country diversity (Maddi & Gingras 2021), institutional diversity, year and disciplines in the regression model to control for confounding factors (see Table 6).

Table 6 The description of control variables

| Variable | Description |
|---|---|
| Team size | The number of authors in the article teams. |
| International team | Following Kuan et al. (2024), we defined international team as teams with authors from two or more distinct countries. This variable is coded as a binary variable ($x = 1$ for international teams, and $x = 0$ for domestic teams). |
| Country diversity | The number of unique countries involved in the teams. |
| Institutional diversity | The number of unique institutions involved in the teams. |
| Year | The publication year of the articles. |
| Discipline | The discipline of the articles. |

Since PLOS does not disambiguate authors' affiliations, we used the DOI of each PLOS article to match it with the corresponding record in the OpenAlex dataset[9]. OpenAlex is a widely used bibliometric dataset that links author affiliation to the Research Organization Registry[10], which provides standardized identifiers and metadata for research organizations worldwide. This procedure allowed us to gather the affiliations and country for each author in every article.

---

[9] https://docs.openalex.org/
[10] https://ror.org/

The empirical model is defined as follows:
$$teamImpact_i = \beta_0 + \beta_1 GDleader_i + \beta_2 GDleader_i^2 + \beta_3 GDsupport_i + \beta_4 GDsupport_i^2 + \beta_5 X_i + D_i + Y_i + \varepsilon \qquad (2)$$

Where $teamImpact_i$ denotes the five-year citation count of article $i$. Following the suggestion of Thelwall and Wilson (2014), we transformed citation counts into logarithmic form prior to regression analysis. To address the issue of zero values rendering the logarithm undefined, we add one to original citation counts before applying the transformation. For article $i$, $GDleader_i$ represents the gender diversity within its leadership group, whereas $GDsupport_i$ denotes that of its support group. Most prior studies conceptualized and empirically examined the relationship between gender diversity and team performance as linear(Yang et al. 2022). Such a specification may overlook the multifaceted effects of gender diversity on team performance, and several scholars have therefore advocated the use of nonlinear models to more accurately capture these effects (Xie et al. 2020; Wullum Nielsen & Börjeson 2019). Accordingly, we include quadratic terms for gender diversity in the leadership group (i.e., $GDleader_i^2$) and in the support group (i.e., $GDsupport_i^2$). Moreover, $X_i$ represents control variables for article $i$. We also included disciplines dummies ($D_i$) and publication year dummies ($Y_i$) to account for disciplinary and temporal effects. Finally, all regressions were estimated using ordinary least squares (OLS).

The descriptive statistics of our variables are presented in Tables A1. We also calculated spearman correlations among the variables, as reported in Table A2. Although some correlations exceed 0.5, the average variance inflation factor (VIF) across all variables is 3.19, which is well below the conventional threshold of 10, indicating that multicollinearity is not a serious concern (Vestal & Danneels 2024).

**(2) Threshold regression**

To answer RQ2, we adopted threshold regression in our study (Hansen 2000). This method is particularly effective for capturing nonlinear relationships between independent and dependent variables. Many variables in empirical research exhibit structural breaks, where predictors cross critical thresholds and induce abrupt changes in their relationship with the dependent variable. In such cases, the threshold effect identifies the transition point as the threshold value. The number of thresholds and their corresponding threshold values can be determined automatically by the model using information criteria such as the Bayesian Information Criterion (BIC), Akaike Information Criterion (AIC), or Hannan-Quinn Information Criterion (HQIC), or they can be specified manually based on domain expertise (Hansen 2000). In this study, both the number of thresholds and their threshold values were determined automatically by the threshold regression models.

Unlike conventional approaches that examine moderating effects through interaction terms in multivariable regression models, threshold regression employs a segmented estimation strategy. It divides the sample into distinct regimes based on identified threshold values and estimates separate regression models within each subsample. This approach offers two key advantages: it provides an intuitive visualization of how variable relationships evolve across different threshold intervals, and it improves the interpretability of results. by contrast, although interaction terms can capture variable interplay, they generally assume that the effects of moderator variables change monotonically, which may not reflect real situation (Hansen 1999).

The threshold regression used in this study is shown as follows:

$$teamImpact_i = \begin{cases} \beta_0 + \beta_1 GDleader_i + \beta_2 GDsupport_i + \beta_3 X_i + D_i + Y_i + \varepsilon, teamSize \leq \gamma \\ \beta_0 + \beta_4 GDleader_i + \beta_5 GDsupport_i + \beta_6 X_i + D_i + Y_i + \varepsilon, teamSize > \gamma \end{cases} \quad (3)$$

Where $teamSize$ is the threshold variable, and $\gamma$ represents the threshold parameter. Threshold regression can be expressed as a piecewise function, when $teamSize \leq \gamma$, the coefficients for $GDleader_i$ and $GDsupport_i$ are $\beta_1$ and $\beta_2$, respectively; when $teamSize > \gamma$, the corresponding coefficients become $\beta_4$ and $\beta_5$. The analysis was conducted using the *threshold* command in STATA (version 17).

# 5. Results

In Section 5.1, we analyzed the relationship between role-specific gender diversity and team impact from an exploratory perspective (RQ1). In section 5.2, we adopted regression analysis to further quantify the relationship between role-specific gender diversity and team impact (RQ1). In Section 5.3, we addressed RQ2 by examining the moderating effect of team size on this relationship. Finally, in Section 5.4, we conducted robustness checks to validate our findings.

## 5.1 The gender diversity of different team roles and team impact: an exploratory perspective

Figure 3A shows the average team impact across the nine team types, and Figure 3B presents the Welch's t-tests pairwise differences, where darker blue indicates higher statistical significance and deeper red indicates lower significance. Teams with male leadership and mixed-gender support (MLead-XSup) achieve the highest average citations (avg. five-year citations = 22.29) among all team types, significantly exceeding other configurations except female-leadership with male support (FLead-MSup). In contrast, teams with female leadership and female support authors (FLead-FSup) exhibit the lowest average impact (avg. five-year citations = 15.70), significantly below all other types (Figure 3B). Male-lead teams with exclusively male support also perform poorly (MLead-MSup), ranking third from the bottom (avg. five-year citations = 17.97). By comparison, mixed-gender leadership and support teams (XLead-XSup) rank second in terms of five-year citations (avg. five-year citations = 21.31). Taken together, these results suggest that gender-diverse team structures tend to enhance team impact, although the magnitude of this advantage depends on the distribution of leadership and support roles.

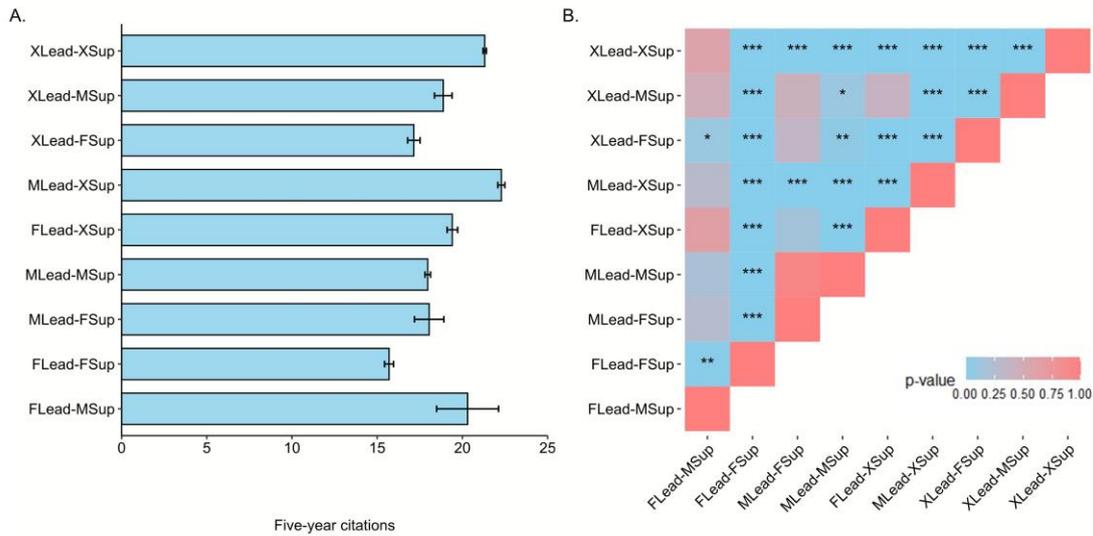

Figure 3 Team impact on different types of teams (Note: *** p < 0.01, ** p < 0.05, * p < 0.1)

Disciplinary differences exist in the gender distribution (Huang et al. 2020), team structure (Xu, Wu, et al. 2022), research paradigms (Jones 2009) and other aspects. Hence, it is necessary to investigate how these patterns vary across disciplines. To this end, we conducted three steps. First, we calculated the average five-year citations for the nine types of teams within each discipline. Second, we ranked the nine types of teams within each discipline based on their average five-year citations. Third, we assessed whether the ranking of the nine types of teams was consistent across disciplines. As shown in Figure 4, the *x*-axis represents three disciplines with the largest sample size that are sorted from left to right in terms of the number of publications, and the *y*-axis represents the ranking of the nine types of teams[11].

Figure 4 shows that teams with male leadership and mixed-gender support (MLead-XSup) rank first or third across the three disciplines, whereas teams with gender-diverse leadership and support (XLead-XSup) consistently rank second or third. Teams with mixed-gender leadership and female support (XLead-FSup) or female leadership and female support (FLead-FSup) occupy positions between last and third from the bottom, which is broadly consistent with the results in Figure 3. Interestingly, male-led teams with male support (MLead-MSup) rank low in Clinical Medicine and Biomedical Research but relatively high in Biology. Overall, disciplinary differences are evident in the impact rankings of the nine team types, and the detailed patterns are not entirely consistent with the aggregate findings in Figure 3. These results further suggest that disciplinary factors should be taken into account in the subsequent regression analysis.

---

[11] Because articles from Clinical Medicine, Biomedical Research, and Biology constitute approximately 80% of our dataset, and the other disciplines are insufficiently represented, we limit our comparison to these three fields.

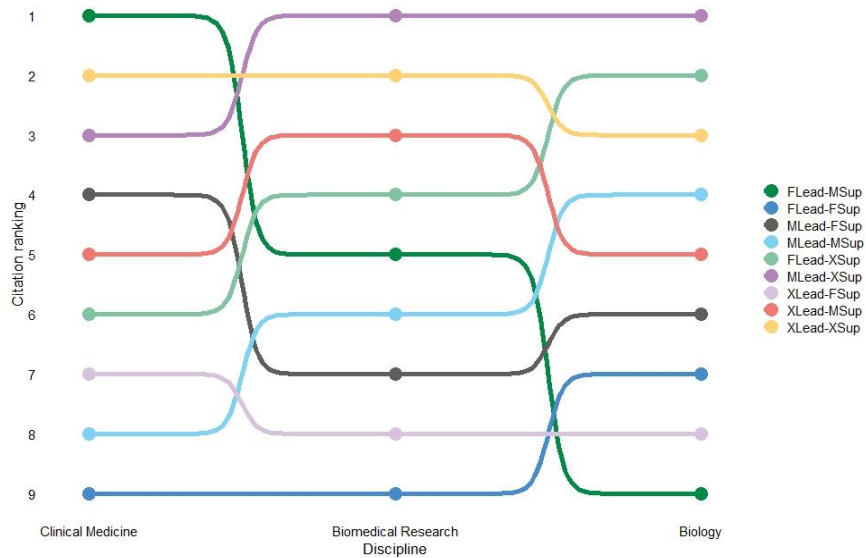

Figure 4 Team impact of nine types of teams by discipline

## 5.2 The gender diversity of different team roles and team impact: the perspective of regression analysis

To confirm our earlier observations, we controlled for potential confounding factors and conducted regression analyses to address RQ1. The regression results are reported in Table 7. Model (1) includes only the control variables. Models (2)-(6) examine the effect of gender diversity on team impact for both leadership and support groups. At the overall team level, without separating leadership and support roles, Model (2) shows a positive and statistically significant association between overall team gender diversity and team impact ($\beta = 0.0603, p < 0.01$). To assess the effect of leadership gender diversity, we introduce both the leadership-group gender diversity (GDleader) and its squared term (GDleader$^2$) in Models (3)-(4). The results indicate that the coefficient for leadership-group gender diversity is significantly positive ($\beta = 0.1746, p < 0.01$), while its squared term is significantly negative ($\beta = -0.1704, p < 0.01$), suggesting an inverted U-shaped relationship between leadership-group gender diversity and team impact. The turning point, where citation impact reaches its maximum, occurs at a gender diversity of approximately $0.51^{12}$ in the leadership group. Similarly, Model (6) includes both the linear and quadratic terms of gender diversity in the support group. The results show that both coefficients are statically significant, with the quadratic term being negative ($\beta = -0.1789, p < 0.01$), again indicating an inverted U-shaped relationship between support-group gender diversity and team impact. The turning point occurs at a gender diversity of approximately $0.64^{13}$. The estimated relationships for both roles are visualized in Figure 5.

---

[12] $\text{Turning point} = -\frac{\text{Coefficient of linear term for GDleader}}{2*\text{Coefficient of quadratic term for GDleader}^2} = -\frac{0.1746}{2*(-0.1704)} \approx 0.51$

[13] $\text{Turning point} = -\frac{\text{Coefficient of linear term for GDsupport}}{2*\text{Coefficient of quadratic term for GDsupport}^2} = -\frac{0.2287}{2*(-0.1789)} \approx 0.64$

Next, both leadership-group gender diversity and support-group gender diversity are included in Model (7). Interestingly, compared with Model (3) and (5), Model (7) shows that the coefficient for leadership-group gender diversity (GDleader) turns significantly negative ($\beta = -0.0231$, $p < 0.01$), whereas the coefficient for support-group gender diversity (GDsupport) remains significantly positive ($\beta = -0.0709$, $p < 0.01$). This pattern suggests that leadership-group gender diversity may exert mixed effects on team impact. On the one hand, gender-diverse teams can provide a broader range of perspectives and solutions that accelerate problem-solving (Yang et al. 2022; Maddi & Gingras 2021). On the other hand, differences in communication styles and experience may create challenges in coordination, thereby reducing team impact (Nielsen et al. 2018).

We then compared the effects of the nine team types in Model (8). Detailed classification of the team types is provided in Table 5, and teams composed of mixed-gender leadership and mixed-gender support groups (XLead-XSup) serve as the reference category. The results reveal notable differences across gender configurations of leadership and support roles. Teams with mixed-gender leadership and male support groups (XLead-MSup) exhibit significantly lower impact than the reference group ($\beta = -0.0412$, $p < 0.01$). Similarly, teams composed of all-male leadership and all-male support (MLead-MSup) and those with all-female leadership and all-female support (FLead-FSup) also receive fewer citations ($\beta = -0.0654$, $p < 0.01; \beta = -0.0353$, $p < 0.01$, respectively). These findings suggest that gender-homogeneous structures, whether male or female, are associated with lower team impact relative to mixed-gender configurations.

In contrast, several heterogeneous configurations show positive and significant effects compared to reference groups. Teams with male leadership and mixed-gender support (MLead-XSup) achieve higher citation counts than the reference group ($\beta = 0.0257$, $p < 0.01$). Notably, teams with female leadership and male support (FLead-MSup) exhibit the highest level of impact among these team compositions ($\beta = 0.1800$, $p < 0.01$).

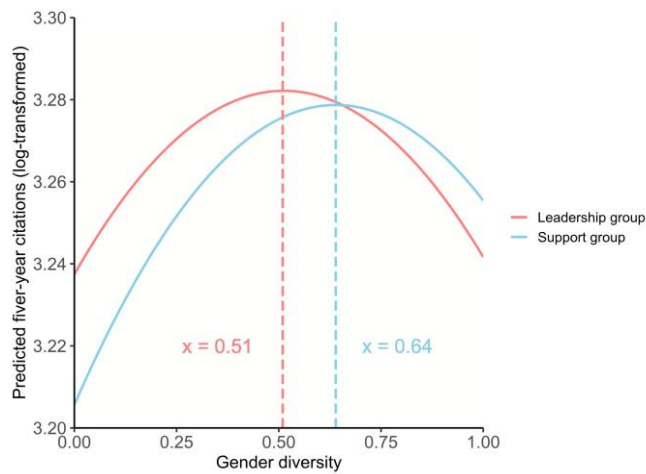

Figure 5 Estimated relationship between gender diversity and team performance for the leadership and support groups

Table 7 The estimates for the influence of gender diversity of leadership and support groups on team impact

| Variables | (1) Five-year citations | (2) Five-year citations | (3) Five-year citations | (4) Five-year citations | (5) Five-year citations | (6) Five-year citations | (7) Five-year citations | (8) Five-year citations |
|---|---|---|---|---|---|---|---|---|
| Gender diversity | | 0.0603*** (0.0065) | | | | | | |
| GDleader | | | 0.0133*** (0.0055) | 0.1746*** (0.0344) | | | -0.0231*** (0.0066) | |
| GDleader$^2$ | | | | -0.1704*** (0.0359) | | | | |
| GDsupport | | | | | 0.0558*** (0.0061) | 0.2287*** (0.0314) | 0.0709*** (0.0075) | |
| GDsupport$^2$ | | | | | | -0.1789*** (0.0319) | | |
| XLead-MSup | | | | | | | | -0.0412** (0.0167) |
| XLead-FSup | | | | | | | | -0.0298* (0.0167) |
| MLead-XSup | | | | | | | | 0.0257*** (0.0075) |
| FLead-XSup | | | | | | | | 0.0273** (0.0135) |
| MLead-MSup | | | | | | | | -0.0654*** (0.0067) |
| MLead-FSup | | | | | | | | 0.0970** (0.0432) |
| FLead-FSup | | | | | | | | -0.0353** (0.0144) |
| FLead-MSup | | | | | | | | 0.1800*** (0.0587) |
| Team size | 0.0362*** (0.0007) | 0.0346*** (0.0007) | 0.0359*** (0.0007) | 0.0354*** (0.0008) | 0.0345*** (0.0007) | 0.0339*** (0.0007) | 0.0345*** (0.0007) | 0.0342*** (0.0007) |
| International team | 0.0372*** (0.0042) | 0.0344*** (0.0071) | 0.0367*** (0.0071) | 0.0350*** (0.0071) | 0.0343*** (0.0071) | 0.0327*** (0.0071) | 0.0346*** (0.0071) | 0.0336*** (0.0071) |
| Country diversity | 0.0128*** (0.0042) | 0.0137*** (0.0042) | 0.0129*** (0.0041) | 0.0133*** (0.0042) | 0.0136*** (0.0042) | 0.0143*** (0.0042) | 0.0138*** (0.0041) | 0.0145*** (0.0042) |
| Institutional diversity | 0.0018 (0.0017) | 0.0025 (0.0017) | 0.0020 (0.0017) | 0.0020 (0.0017) | 0.0027 (0.0017) | 0.0028 (0.0017) | 0.0026 (0.0017) | 0.0026 (0.0017) |
| Constant | 2.9877*** (0.0198) | 2.9554*** (0.0201) | 2.9825*** (0.0198) | 2.9823*** (0.0198) | 2.9613*** (0.0199) | 2.9568 (0.0200) | 2.9632*** (0.0200) | 3.0095*** (0.0200) |
| Discipline fixed effect | Yes | Yes | Yes | Yes | Yes | Yes | Yes | Yes |
| Year fixed effect | Yes | Yes | Yes | Yes | Yes | Yes | Yes | Yes |
| Observations | 133,858 | 133,858 | 133,858 | 133,858 | 133,858 | 133,858 | 133,858 | 133,858 |
| Adj $R^2$ | 0.1285 | 0.1292 | 0.1286 | 0.1287 | 0.1292 | 0.1292 | 0.1291 | 0.1295 |

**Notes**: Standard errors in parentheses. *** $p < 0.01$, ** $p < 0.05$, * $p < 0.1$. All 15 disciplinary categories listed in Table 4 are included in the models.

## 5.3 The moderating role of team size

Table 8 The relationship between gender diversity of different team roles and team impact depends on team size

| Variable | (1) Five-year citations (team size<=6) | Five-year citations (team size >6) | (2) Five-year citations (team size <=6) | (3) Five-year citations (team size >6) |
|---|---|---|---|---|
| Gender diversity | | | 0.0261*** | 0.0361** |
| | | | (0.0073) | (0.0164) |
| GDleader | -0.0396*** | 0.0032 | | |
| | (0.0081) | (0.0118) | | |
| GDsupport | 0.0432*** | 0.0393** | | |
| | (0.0087) | (0.0176) | | |
| Team size | 0.0555*** | 0.0219*** | 0.0541*** | 0.0219*** |
| | (0.0023) | (0.0011) | (0.0023) | (0.0010) |
| International team | 0.0111 | 0.0269** | 0.0109 | 0.0271*** |
| | (0.0136) | (0.0106) | (0.0136) | (0.0102) |
| Country diversity | -0.0008 | 0.0266*** | -0.0010 | 0.0266*** |
| | (0.0101) | (0.0048) | (0.0101) | (0.0046) |
| Institutional diversity | 0.0116*** | 0.0013 | 0.0120*** | 0.0012 |
| | (0.0029) | (0.0022) | (0.0029) | (0.0020) |
| Constant | 2.8574*** | 3.1987*** | 2.8510*** | 3.2024*** |
| | (0.0270) | (0.0366) | (0.0270) | (0.0357) |
| Discipline fixed effect | Yes | Yes | Yes | Yes |
| Year fixed effect | Yes | Yes | Yes | Yes |
| Observations | 133,858 | | 86,565 | 47,293 |
| Adj R² | | | 0.1073 | 0.0981 |
| BIC | 0.0004 | | | |

Notes: Standard errors in parentheses. *** $p < 0.01$, ** $p < 0.05$, * $p < 0.1$. All 15 disciplinary categories listed in Table 4 are included in the models.

We used threshold regression to explore the moderating effect of team size (i.e., answering RQ2). Using STATA 17, the model identified a single threshold at a team size of six (see Table 8, Model (1)). When team size is six or fewer, leadership-group gender diversity is significantly and negatively associated with team impact ($\beta = -0.0396, p < 0.01$), whereas support-group gender diversity shows a significant positive effect ($\beta = 0.0432, p < 0.01$). For team sizes larger than six, the coefficient for leadership-group gender diversity turns positive but becomes statistically insignificant ($\beta = 0.0032, p > 0.1$), whereas support-group gender remains significantly and positively associated with team impact ($\beta = 0.0393, p < 0.05$).

These findings suggest that team size moderates the relationship between leadership-group gender diversity and team impact. In small teams, gender-diverse leadership diminishes team impact, whereas in larger teams, it begins to exert positive effects, although the results are not statistically significant.

Moreover, it is important to examine whether team size also moderates the association between overall team gender diversity and team impact. Based on the estimated threshold, we categorize teams into two groups: Group 1 includes teams with six or fewer members, and Group 2 includes teams with more than six members. We run separate regressions for each group. Models (2) and (3) show that overall team gender diversity is significantly and positively related to team impact. These results highlight the imperative need to account for both team roles and team size when exploring the potential link between team gender diversity and team performance.

### 5.4 Robustness check

We also conducted robustness tests to ensure the reliability of our conclusions. First, we re-estimated the model using different citation windows to measure team impact. Table A3 and Table A4 present the results based on three-year and four-year citations, respectively. The main findings remain consistent with the results reported in Table 7. Second, as shown in Table 3, about 10% of the articles were not assigned to any discipline. These articles were classified into a single category (i.e., other discipline), and were included in the regression analysis. Moreover, the disciplinary distribution of our dataset is unbalanced. To test robustness, we restricted the sample to the three disciplines with the largest representation, including Clinical Medicine, Biomedical Research, and Biology. Compared with the results from the overall sample, the coefficients reported in Table A5 show that both leadership-group and support-group gender diversity have inverted U-shaped relationships with team impact, and that teams composed of female leadership and male support achieve the highest impact among all team types, which is consistent with the results reported in Table 7. Third, since the dependent variable is a count measure and indicates over-dispersion (mean = 20.46, S.D. = 26.90), we adopted negative binomial regression instead of OLS to estimate the models. As shown in Table A46, the key findings remain robust, except that the coefficients for some team types (e.g., XLead-MSup and XLead-FSup) become insignificant. Nevertheless, the results still demonstrate that gender diversity produces positive team impacts only when it is reflected across different roles. Taken together, these analyses indicate that the empirical findings of this study are robust.

# 6. Discussion

Gender as an social identity category that can shape thoughts, perspectives, experience and perceptions (Lindsey 2015). Gender-diverse teams demonstrate a greater capacity to integrate varied experiences, beliefs, and resources structured along gendered lines, thereby exhibiting enhanced potential for generating novel and impactful outcomes (Wullum Nielsen & Börjeson 2019). Beyond confirming this insight, our study extends existing knowledge by examining how gender diversity relates to team performance through inclusion of the perspective of team roles. Although previous studies often assume gender diversity is universally beneficial, our findings reveal a more nuanced effect that depends on team role composition. This underscores the importance of moving beyond a one-size-fits-all perspective and considering how gender dynamics and team roles jointly shape collaboration, communication and performance. These insights can inform the best practices for optimizing research team construction and management.

Our empirical findings reveal an inverted-U-shaped relationship between gender diversity and team impact for both leadership and support roles, which is consistent with the categorization-elaboration model (CEM) (van Knippenberg et al. 2004). According to this theory, various forms of team diversity (e.g., gender diversity, age diversity) can simultaneously trigger subgroup categorization processes (Tajfel 1982) and information-processing mechanisms, leading to mixed effects on team performance. In our context, at lower level, greater gender diversity for both team roles are associated with better team impact. A possible explanation is that gender-diverse teams draw on a broader range of skills and experiences, thereby facilitating discussion and enhancing performance. However, once gender diversity surpasses the "sweet spot" (0.51 for support groups and 0.64 for support groups), further increases are linked to lower team impact. At this stage, intergroup divisions may emerge from a more diverse and balanced gender composition, impeding communication, interaction, and collaboration among team members, and ultimately reducing team impact. These empirical findings provide strong support for the CEM.

In this study, scientific teams are categorized into nine types based on gender composition across leadership and support roles. Teams with all-female leadership and all-male support achieve the highest impact, whereas teams with all-male leadership and all-male support exhibit the lowest impact. These findings highlight that gender diversity influences team performance in role-specific ways: what matters is not merely the aggregate proportion of male and female members, but how gender composition aligns with hierarchical role differentiation. Depending on their configuration, leadership and support positions may either amplify or mitigate the benefits of gender diversity.

We also find that leadership-group gender diversity may diminish team impact in small teams, whereas in large teams it shows a positive, though statistically nonsignificant. Small teams typically have flatter power structures and simpler interpersonal dynamics (Xu, Bu, et al. 2022). As a result, negative effects arising from intergroup bias are more pronounced, leading leadership gender diversity to hinder team performance. By contrast, in large teams, gender-mixed leadership groups are better able to manage the challenges of diversity that accompany increased team sizes. Under such conditions, information elaboration is more likely to occur, thereby promoting the team's potential to produce high-impact outcomes (Homan et al. 2020).

## 6.1 Theoretical and practical implications

This study also has theoretical and practical implications. From a theoretical perspective, these findings extend research on gender diversity and team performance by highlighting the importance of role differentiation within teams. Previous studies generally operationalized diversity in aggregate terms, focusing on overall gender diversity(Yang et al. 2022; Zhang et al. 2025). Our findings suggest that such an approach may overlook critical role-based mechanisms. Specifically, the gender composition of leadership and support groups plays a critical role in shaping team impact. This underscores the need to refine theories of diversity (e.g., CEM) by incorporating team roles, rather than treating diversity as a uniform construct.

Practically, these insights carry important implications for funding allocation and team design. Funding agencies aiming to maximize research impact should pay attention not only to gender balance at the aggregate level of applicant teams but also to the distribution of gender diversity across leadership and support roles. For scientific team construction, principal investigators (PIs) of large teams may benefit from diversifying their leadership groups by gender, whereas PIs of small teams may achieve greater impact by diversifying their support groups. Deliberately considering gender-role composition in team design represents a viable strategy for enhancing the academic influence of scientific teams.

## 6.2 Limitations

Although this study conducted a large-scale quantitive analysis of the effect of gender diversity across team roles on team impact, there are still several limitations in this study.

First, due to the limitations of Genderize.io, most of the Chinese and Korean authors were excluded from our dataset. Including these authors in future research would yield a more comprehensive results. Second, Genderize.io is a binary gender inference system, which prevents accurate identification of individuals with non-binary identities. Developing new tools that incorporate diverse gender identities would help to address this shortcoming in future work. Third, although the PLOS dataset is widely used for conducting science of science research, most of the articles in this dataset come from biomedical-related disicplines. Therefore, the results should be interpreted with caution before being generalized to other disciplines. In the future, we plan to expand the dataset to include articles from a broader range of disciplines. This will be possible once author contribution statements become available in other datasets, allowing us to further validate the generalizability of our findings. Fourth, citation counts were used in our study to characterize team impact. However, citations represents only one dimension of impact. Future research could investigate the relationship between role-specific gender diversity and other indicators of team success, such as productivity, disruptive innovation (Zhao et al. 2024), technological impact, and combinational novelty (Liu et al. 2024), which would be of considerable interest to academia. Fifth, although the role assignment method applied was adapted from a study published in a prestigious journal (Xu, Wu, et al. 2022), its appropriateness for the new version of the contribution statement taxonomy remains uncertain. A more generalized method for role assignment should be developed in future studies. Sixth, in our study, each paper was assigned to a single discipline, which may not fully capture the interdisciplinary nature of certain works and could introduce potential bias.

# 7. Conclusion

Grounded in the question of whether team gender heterogeneity is invariably beneficial for team impact, this study examined the influence of role-specific gender diversity on scientific impact, and the moderating role of team size. Drawing on 161,193 research articles published in PLOS journals, the results reveal that (1) there exist an inverted U-shaped relationship between gender diversity and team impact for both leadership and support roles; (2) female-led teams with male support yield highest team impact compared among all team types;(3) leadership-group gender diversity is significantly and negatively associated with team impact when team size is six or fewer, but the relationship becomes positive once team size exceeds six. These insights advance the literature on gender diversity and teams by unpacking the complex interplay between gender diversity, team roles, team size, and performance.

# Acknowledgements

Yi Zhao would like to thank Dr. Tongbin Yang from Jiangsu University for his valuable suggestions. This work was supported by the National Research Foundation of Korea (NRF) (No. RS-2023-00209775), ICONS (Institute of Convergence Science), Yonsei University, the National Natural Science Foundation of China (No. 72074113, No. 72004054), and National Social Science Fund of China (No. 24CTQ027).

# Appendix

Table A1 Descriptive statistics

| Variable | N | Mean | SD | Min | Median | Max |
|---|---|---|---|---|---|---|
| Five-year citations | 133858 | 20.46 | 26.898 | 0 | 14 | 1538 |
| Gender diversity | 133858 | 0.70 | 0.376 | 0 | 0.890 | 1 |
| GDleader | 133858 | 0.57 | 0.439 | 0 | 0.811 | 1 |
| GDsupport | 133858 | 0.66 | 0.401 | 0 | 0.863 | 1 |
| Team size | 133858 | 6.07 | 4.191 | 2 | 5 | 346 |
| International team | 133858 | 0.39 | 0.487 | 0 | 0 | 1 |
| Country diversity | 133858 | 1.59 | 1.003 | 1 | 1 | 24 |
| Institutional diversity | 133858 | 2.84 | 2.164 | 1 | 2 | 62 |

Table A2 Spearman correlation matrix

| No. | Variable | VIF | 1 | 2 | 3 | 4 | 5 | 6 | 7 | 8 |
|---|---|---|---|---|---|---|---|---|---|---|
| 1 | Five-year citations | | 1 | | | | | | | |
| 2 | Gender diversity | 5.88 | 0.044 | 1 | | | | | | |
| 3 | GDleader | 4.65 | 0.023 | 0.710 | 1 | | | | | |
| 4 | GDsupport | 3.22 | 0.050 | 0.883 | 0.601 | 1 | | | | |
| 5 | Team size | 1.83 | 0.171 | 0.264 | 0.202 | 0.292 | 1 | | | |
| 6 | International team | 2.21 | 0.054 | 0.067 | 0.066 | 0.071 | 0.227 | 1 | | |
| 7 | Country diversity | 3.22 | 0.102 | 0.082 | 0.081 | 0.090 | 0.416 | 0.734 | 1 | |
| 8 | Institutional diversity | 2.52 | 0.128 | 0.125 | 0.103 | 0.134 | 0.636 | 0.435 | 0.668 | 1 |

Note: all coefficients are significant at the 99% level of significance (P<0.01).

Table A3 Effects of gender diversity in leadership and support groups on team impact (measured by three-year citations)

| Variables | (1) Three-year citations | (2) Three-year citations | (3) Three-year citations | (4) Three-year citations |
|---|---|---|---|---|
| GDleader | 0.145*** (0.033) | | -0.026*** (0.006) | |
| GDleader$^2$ | -0.145*** (0.035) | | | |
| GDsupport | | 0.203*** (0.030) | 0.066*** (0.007) | |
| GDsupport$^2$ | | -0.159*** (0.031) | | |
| XLead-MSup | | | | -0.026 (0.016) |
| XLead-FSup | | | | -0.033** (0.016) |
| MLead-XSup | | | | 0.032*** (0.007) |
| FLead-XSup | | | | 0.024* (0.013) |
| MLead-MSup | | | | -0.055*** (0.006) |
| MLead-FSup | | | | 0.098** (0.042) |
| FLead-FSup | | | | -0.044*** (0.014) |
| FLead-MSup | | | | 0.179*** (0.057) |
| Team size | 0.035*** (0.001) | 0.034*** (0.001) | 0.034*** (0.001) | 0.034*** (0.001) |
| International team | 0.031*** (0.007) | 0.029*** (0.007) | 0.031*** (0.007) | 0.029*** (0.007) |
| Country diversity | 0.014*** (0.004) | 0.015*** (0.004) | 0.014*** (0.004) | 0.015*** (0.004) |
| Institutional diversity | 0.001 (0.002) | 0.002 (0.002) | 0.002 (0.002) | 0.002 (0.002) |
| Constant | 2.475*** (0.019) | 2.451*** (0.019) | 2.457*** (0.019) | 2.495*** (0.019) |
| Discipline fixed effect | Yes | Yes | Yes | Yes |
| Year fixed effect | Yes | Yes | Yes | Yes |
| Observations | 133,858 | 133,858 | 133,858 | 133,858 |
| Adj R-squared | 0.1315 | 0.1320 | 0.1320 | 0.1323 |

**Notes**: Standard errors in parentheses. *** $p < 0.01$, ** $p < 0.05$, * $p < 0.1$.

Table A4 Effects of gender diversity in leadership and support groups on team impact (measured by four-year citations)

| Variables | (1) Four-year citations | (2) Four-year citations | (3) Four-year citations | (4) Four-year citations |
|---|---|---|---|---|
| GDleader | 0.156*** (0.034) | | -0.025*** (0.007) | |
| GDleader$^2$ | -0.154*** (0.035) | | | |
| GDsupport | | 0.212*** (0.031) | 0.070*** (0.007) | |
| GDsupport$^2$ | | -0.164*** (0.031) | | |
| XLead-MSup | | | | -0.035** (0.016) |
| XLead-FSup | | | | -0.037** (0.016) |
| MLead-XSup | | | | 0.028*** (0.007) |
| FLead-XSup | | | | 0.027** (0.013) |
| MLead-MSup | | | | -0.060*** (0.007) |
| MLead-FSup | | | | 0.096** (0.042) |
| FLead-FSup | | | | -0.040*** (0.014) |
| FLead-MSup | | | | 0.157*** (0.058) |
| Team size | 0.035*** (0.001) | 0.034*** (0.001) | 0.035*** (0.001) | 0.034*** (0.001) |
| International team | 0.035*** (0.007) | 0.033*** (0.007) | 0.034*** (0.007) | 0.033*** (0.007) |
| Country diversity | 0.013*** (0.004) | 0.014*** (0.004) | 0.013*** (0.004) | 0.014*** (0.004) |
| Institutional diversity | 0.002 (0.002) | 0.003* (0.002) | 0.003 (0.002) | 0.003 (0.002) |
| Constant | 2.760*** (0.020) | 2.735*** (0.020) | 2.741*** (0.020) | 2.784*** (0.020) |
| Discipline fixed effect | Yes | Yes | Yes | Yes |
| Year fixed effect | Yes | Yes | Yes | Yes |
| Observations | 133,858 | 133,858 | 133,858 | 133,858 |
| Adj R-squared | 0.1331 | 0.1336 | 0.1336 | 0.1339 |

**Notes**: Standard errors in parentheses. *** $p < 0.01$, ** $p < 0.05$, * $p < 0.1$.

Table A5 Effects of gender diversity in leadership and support groups on team impact (sample restricted to Clinical Medicine, Biomedical Research, and Biology)

| Variables | (1) Five-year citations | (2) Five-year citations | (3) Five-year citations | (4) Five-year citations |
|---|---|---|---|---|
| GDleader | 0.131*** (0.037) | | -0.024*** (0.007) | |
| GDleader$^2$ | -0.131*** (0.039) | | | |
| GDsupport | | 0.176*** (0.034) | 0.063*** (0.008) | |
| GDsupport$^2$ | | -0.132*** (0.034) | | |
| XLead-MSup | | | | -0.034* (0.019) |
| XLead-FSup | | | | -0.026 (0.019) |
| MLead-XSup | | | | 0.023*** (0.008) |
| FLead-XSup | | | | 0.029** (0.014) |
| MLead-MSup | | | | -0.055*** (0.007) |
| MLead-FSup | | | | 0.107** (0.046) |
| FLead-FSup | | | | -0.032* (0.016) |
| FLead-MSup | | | | 0.225*** (0.063) |
| Team size | 0.034*** (0.001) | 0.032*** (0.001) | 0.033*** (0.001) | 0.033*** (0.001) |
| International team | 0.031*** (0.008) | 0.029*** (0.008) | 0.031*** (0.008) | 0.030*** (0.008) |
| Country diversity | 0.016*** (0.005) | 0.016*** (0.005) | 0.016*** (0.005) | 0.017*** (0.005) |
| Institutional diversity | 0.003 (0.002) | 0.004** (0.002) | 0.003* (0.002) | 0.003* (0.002) |
| Constant | 5.244*** (0.843) | 5.248*** (0.843) | 5.236*** (0.843) | 5.290*** (0.842) |
| Discipline fixed effect | Yes | Yes | Yes | Yes |
| Year fixed effect | Yes | Yes | Yes | Yes |
| Observations | 108,531 | 108,531 | 108,531 | 108,531 |
| Adj R-squared | 0.0975 | 0.0979 | 0.0978 | 0.0982 |

**Notes**: Standard errors in parentheses. *** $p < 0.01$, ** $p < 0.05$, * $p < 0.1$.

Table A6 Effects of gender diversity in leadership and support groups on team impact (Negative Binomial regression estimates)

| Variables | (1) Five-year citations | (2) Five-year citations | (3) Five-year citations | (4) Five-year citations |
|---|---|---|---|---|
| GDleader | 0.165*** (0.035) | | -0.017** (0.007) | |
| GDleader$^2$ | -0.179*** (0.036) | | | |
| GDsupport | | 0.148*** (0.032) | 0.023*** (0.008) | |
| GDsupport$^2$ | | -0.140*** (0.033) | | |
| XLead-MSup | | | | 0.001 (0.017) |
| XLead-FSup | | | | -0.019 (0.017) |
| MLead-XSup | | | | 0.020*** (0.008) |
| FLead-XSup | | | | -0.008 (0.014) |
| MLead-MSup | | | | -0.015** (0.007) |
| MLead-FSup | | | | 0.071 (0.044) |
| FLead-FSup | | | | -0.052*** (0.015) |
| FLead-MSup | | | | 0.201*** (0.060) |
| Team size | 0.038*** (0.001) | 0.037*** (0.001) | 0.038*** (0.001) | 0.038*** (0.001) |
| International team | -0.012* (0.007) | -0.013* (0.007) | -0.011 (0.007) | -0.013* (0.007) |
| Country diversity | 0.028*** (0.004) | 0.029*** (0.004) | 0.028*** (0.004) | 0.029*** (0.004) |
| Institutional diversity | 0.007*** (0.002) | 0.007*** (0.002) | 0.007*** (0.002) | 0.007*** (0.002) |
| Constant | 3.320*** (0.020) | 3.309*** (0.020) | 3.314*** (0.020) | 3.323*** (0.020) |
| lnalpha | -0.347*** (0.004) | -0.347*** (0.004) | -0.347*** (0.004) | -0.347*** (0.004) |
| Discipline fixed effect | Yes | Yes | Yes | Yes |
| Year fixed effect | Yes | Yes | Yes | Yes |
| Observations | 133,858 | 133,858 | 133,858 | 133,858 |
| Log likelihood | -530364.75 | -530366.23 | -530372.45 | -530355.93 |
| Prob > chi$^2$ | 0.0000 | 0.0000 | 0.0000 | 0.0000 |
| Pseudo R$^2$ | 0.0170 | 0.0170 | 0.0170 | 0.0170 |

**Notes**: Standard errors in parentheses. *** $p < 0.01$, ** $p < 0.05$, * $p < 0.1$.